# Optimization of Autonomous Driving Image Detection Based on RFAConv and Triplet Attention

Zhipeng Ling[1*], Qi Xin[1.2], Yiyu Lin[2], Guangze Su[3], Zuwei Shui[4]

[1]Computer Science,University of Sydney,Sydney, Australia
[1.2]Management Information Systems, University of Pittsburgh, Pittsburgh, PA, USA
[2]Computer Science and Engineering, Santa Clara University, CA,USA
[3]Information Studies,Trine University,Phoenix,USA
[4]Information Studies,Trine University,Phoenix,USA

***Corresponding author**:[Zhipeng Ling,E-mail:zlin5578@alumni.sydney.edu.au]

**Abstract.**

YOLOv8 plays a crucial role in the realm of autonomous driving, owing to its high-speed target detection, precise identification and positioning, and versatile compatibility across multiple platforms. By processing video streams or images in real-time, YOLOv8 rapidly and accurately identifies obstacles such as vehicles and pedestrians on roadways, offering essential visual data for autonomous driving systems. Moreover, YOLOv8 supports various tasks including instance segmentation, image classification, and attitude estimation, thereby providing comprehensive visual perception for autonomous driving, ultimately enhancing driving safety and efficiency. Recognizing the significance of object detection in autonomous driving scenarios and the challenges faced by existing methods, this paper proposes a holistic approach to enhance the YOLOv8 model. The study introduces two pivotal modifications: the C2f_RFAConv module and the Triplet Attention mechanism. Firstly, the proposed modifications are elaborated upon in the methodological section. The C2f_RFAConv module replaces the original module to enhance feature extraction efficiency, while the Triplet Attention mechanism enhances feature focus. Subsequently, the experimental procedure delineates the training and evaluation process, encompassing training the original YOLOv8, integrating modified modules, and assessing performance improvements using metrics and PR curves. The results demonstrate the efficacy of the modifications, with the improved YOLOv8 model exhibiting significant performance enhancements, including increased MAP values and improvements in PR curves. Lastly, the analysis section elucidates the results and attributes the performance improvements to the introduced modules. C2f_RFAConv enhances feature extraction efficiency, while Triplet Attention improves feature focus for enhanced target detection.

# 1. Introduction

With the rapid development of autonomous driving technology, the accurate perception and understanding of the surrounding environment of the vehicle has become crucial. Image detection, as the core component of the automatic driving system, directly affects the decision-making and safety performance of the vehicle. In recent years, deep learning, especially convolutional neural networks (CNNs), have made remarkable progress in image detection tasks, such as the [1]YOLO series models (You Only Look Once), which perform well in real-time and detection accuracy. However, faced with the complex and changeable traffic environment, existing methods still face challenges in feature extraction and multi-scale information processing. In practical applications of autonomous driving, image detection needs to deal with a large number of complex scenes, including different lighting conditions, weather changes, and dynamic targets. These factors present the following challenges to detection models:

1. Multi-scale target detection: Targets in the traffic environment vary in size, ranging from distant pedestrians to nearby vehicles, requiring detection models to have strong multi-scale perception abilities.
2. Real-time processing: The automatic driving system requires efficient computing performance to ensure real-time processing and response, thereby avoiding safety hazards caused by delays.
3. Robustness: [2]The model needs to maintain stable performance in various extreme conditions, including rain, night, and occlusion.
4. Accuracy of feature extraction: In complex traffic scenes, accurate extraction of useful features is crucial for target recognition and classification.

Therefore, to address the above challenges, this paper proposes an improved YOLOv8 model that combines the RFAConv and Triplet Attention mechanisms. RFAConv (Receptive Field Attention Convolution) and the Triplet Attention mechanism are introduced to optimize the performance of the YOLOv8 model and improve the image detection capability of the automatic driving system. It is demonstrated that the YOLOv8 model combined with RFAConv and the Triplet Attention mechanism can not only extract and represent key information in images more accurately but also enhance the image detection performance of the automatic driving system while ensuring real-time performance. This paper will detail the design and implementation of these two mechanisms and verify their effectiveness through experiments.

# 2. Related Work

## 2.1. Automatic driving image detection technology

Normally, autopilot uses eight cameras to identify objects in the real world. The images captured by the cameras include pedestrians, vehicles, animals, and obstacles, which are important not only for the safety of the drivers of unmanned autonomous vehicles, but also for others. It is important that the camera is able to identify these objects in a timely and accurate manner.

The following is the automatic driving image recognition framework layer

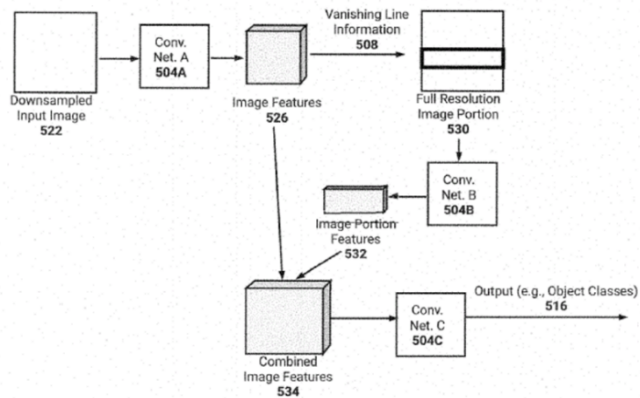

**Figure 1.** Automatic driving image detection

Breakthroughs in deep learning and computer vision have facilitated the rapid development of autonomous driving, which promises to free up drivers, reduce traffic congestion and improve road safety. However, the potential of autonomous driving has yet to be fully unlocked, largely due to unsatisfactory perceptual performance in real-world driving scenarios. As a result, even though autonomous vehicles (AVs)[3] are already being used in many restricted and controlled environments, deploying them in urban environments still faces serious technical challenges. In the real world, performing 3D object detection with a single type of sensor data is not nearly enough. First, each type of sensor data has its own inherent limitations and drawbacks. For example, camera-based systems lack accurate depth information, while liDAR[4] only systems are hampered by low resolution input data, especially over long distances. Table 1 shows this. Overall, for objects with a far ego-sensor (> 60m in KITTI), there are usually fewer than 10 LiDAR points, but the image is more than 400 image pixels. Second, the sensing system must be resistant to sensor failure, failure, or poor performance, so there needs to be more than one type of sensor. Third, data from different sensors naturally complement each other. Their combination allows for a more comprehensive description of the environment, leading to better detection results.

The YOLOv8 model excels in image detection for autonomous driving due to its real-time processing capabilities, high detection accuracy, and robustness in handling complex and dynamic traffic environments. By efficiently processing multi-scale information and accurately extracting features, YOLOv8 enables reliable target recognition and classification, contributing to the safety and efficiency of autonomous driving systems.

*2.2. YOLOv8 model*

The YOLO (You Only Look Once) series of models became very famous in the field of computer vision. YOLO is famous because it has a fairly high accuracy while maintaining the size of a small model. YOLO models can be trained on a single GPU, which makes them suitable for a wide range of developers. Machine learning practitioners can deploy it at low cost on edge hardware or in the cloud. Since it was first released by Joseph Redmond in 2015, YOLO has been on the radar of the computer vision community. In earlier versions (versions 1-4), YOLO was maintained in C code in a custom deep learning framework called Darknet written by Redmond.

The object detection and tracking model YOLOv8 can quickly and accurately identify and locate multiple objects in an image or video frame, as well as track their movement and classify them. In addition to detecting objects, YOLOv8[5] can also distinguish the exact contours of objects, perform instance segmentation, estimate human posture, and help identify and analyze specific patterns in medical images, among other computer vision tasks.

**The main functions of YOLOv8 model include:**

1. High-speed target detection: YOLOv8 continues to maintain the high-speed detection characteristics of YOLO series models, capable of real-time processing of video streams or high-speed analysis of targets in static images.

2. High-precision recognition: Through the improved algorithm and network structure, YOLOv8 improves the accuracy of target detection, including better boundary frame positioning and classification accuracy.

3, multi-platform compatibility: YOLOv8[6] supports deployment through a variety of formats such as ONNX, OpenVINO, CoreML and TFLite, enhancing the availability and compatibility of the model, enabling it to run on a variety of hardware and platforms.

4. Multi-task capability: In addition to target detection, YOLOv8 also supports tasks such as instance segmentation, image classification and pose estimation, providing a one-stop solution for a variety of visual recognition needs.

**Application scenarios of YOLOv8 model:**

Object detection: YOLOv8 is able to quickly and accurately identify and locate multiple objects in an image or video frame. This is particularly useful for security monitoring, traffic flow monitoring, retail analysis and other fields. Instance segmentation: In addition to detecting objects, YOLOv8 can also distinguish the exact contours of objects, which is very important for applications that require accurate object shape information, such as medical image analysis, precision agriculture.

Image classification: YOLOv8 can identify the main content in an image and classify it, which is very useful for applications such as automatic image sorting, content discovery and recommendation systems.

Posture estimation: YOLOv8 can estimate the posture of the human body, which has a wide range of applications in sports analysis, human-computer interaction, motion recognition and other fields.

Tracking: In video, YOLOv8 can not only detect objects, but also track their movement, which is useful for video surveillance, motion analysis, and interactive media production.

Autonomous driving: By accurately identifying and locating vehicles, pedestrians and other obstacles on the road, YOLOv8 can provide vital visual information to autonomous driving systems.

Augmented Reality (AR) : YOLOv8 can recognize objects and scenes in the real world in real time, providing a foundation for AR applications to create richer, interactive user experiences.

In the field of segmented tasks, a milestone in the field is the full convolutional network (FCN). In addition, U-Net and SegNet are common models in drivable area segmentation tasks. However, lane line segmentation is different from driveable area segmentation because lane lines have a distinct elongated feature in road images. The segmentation of lane lines requires effective analysis of Low-Level and multi-scale features.

Recently, models such as PointLaneNet and [7]MFIALane have gained a lot of attention in the field of lane segmentation. Although they achieve excellent results in their respective fields, integrating them together is challenging due to the different resolutions of the required features, with segmentation operations performed at the pixel level and detection tasks using grid cells in a one-stage approach and selective search in a two-stage approach. Despite their different focal points, both segmentation and detection tasks need to extract initial features from the input image. Therefore, a Backbone network can be shared. Integrating three different necks and heads into a model with a shared Backbone saves a lot of computational resources and inference time compared to using separate models for each task.

*2.3. Triplet Attention mechanism principle*

The recent proliferation of attention mechanisms across various computer vision tasks underscores their efficacy in leveraging interdependencies among channels or spatial positions. In this paper, we explore a lightweight yet effective attention mechanism termed Triplet Attention, which captures cross-dimensional interactions using a tripartite structure to compute attention weights. Triplet Attention establishes dependencies between dimensions through a residual transformation following rotational operations on input tensors, encoding information across channels and spatial locations with negligible computational overhead. Our method is straightforward and efficient, seamlessly integrable as an additional module into classic backbone networks. We validate the effectiveness of our approach on challenging tasks including image classification on ImageNet-1k and object detection on [8]MSCOCO and PASCAL VOC datasets. Furthermore, through intuitive examination of GradCAM and GradCAM++ results, we offer deeper insights into the performance of Triplet Attention. Empirical evaluations corroborate our intuition, highlighting the importance of capturing cross-dimensional dependencies when computing attention weights.

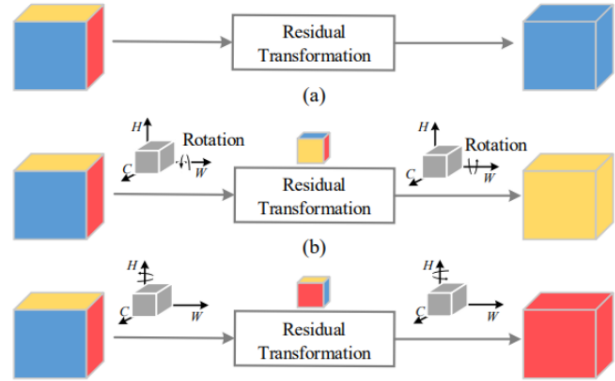

**Figure 2.** Schematic diagram of Triplet Attention

The basic principle of Triplet Attention lies in leveraging a tripartite structure to capture cross-dimensional interactions in input data, thereby computing attention weights. This method effectively establishes interdependencies between channels or spatial positions, with minimal computational cost. Triplet Attention consists of three branches, each responsible for capturing interactions between spatial dimensions (H or W) and channel dimensions. By subjecting the input tensors in each branch to permutation transformations, followed by pooling operations and a convolutional layer of size kxk, attention weights are generated. These weights are then passed through a sigmoid activation layer and applied to the permuted input tensors before being transformed back to their original input shape. This process effectively enhances the model's ability to focus on relevant features across different dimensions, facilitating more effective feature extraction and representation. For example, in image classification tasks, Triplet Attention can help the model to attend to important spatial and channel information simultaneously, improving classification accuracy across various classes and scenarios.

*2.4. Triplet Attention and other simple attention mechanisms*

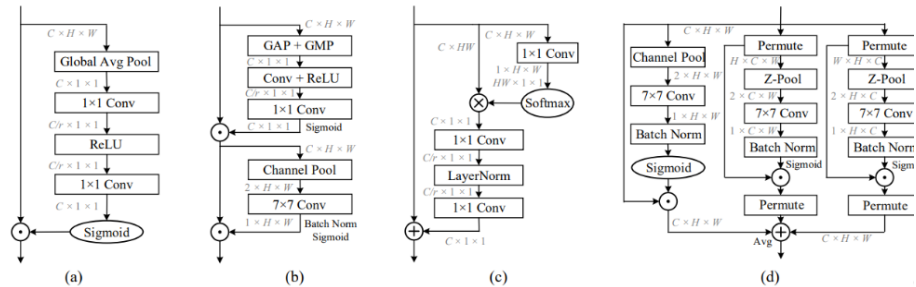

**Figure 3.** Comparison of Triplet Attention with other simple attention mechanisms

In the realm of attention mechanisms for enhancing neural network performance in computer vision tasks, various modules have been proposed. Each module employs distinct strategies to compute attention weights and amplify the significance of salient features within input data. Here, we delineate the key characteristics and operational methodologies of four prominent attention modules, including Squeeze Excitation (SE)[9], Convolutional Block Attention Module (CBAM), Global Context (GC), and our proposed Triplet Attention module:

1. Squeeze Excitation (SE) Module:
- Utilizes global average pooling to generate channel descriptors.
- Employs two fully connected layers (1x1 Conv) with ReLU activation function, followed by a Sigmoid function to produce channel-wise weights.

2. Convolutional Block Attention Module (CBAM):
- Integrates global average pooling and global max pooling (GAP+GMP), followed by convolutional layers and ReLU activation, culminating in a Sigmoid function to compute attention weights.

3. Global Context (GC) Module:
- Initiates with a 1x1 convolutional layer, followed by normalization using Softmax function.
- Subsequently, another 1x1 convolutional layer is applied, followed by LayerNorm and final 1x1 convolution, which are combined with the original feature map using broadcast addition.

4. Triplet Attention :
- Comprises three branches, each dedicated to distinct types of feature interactions.
- The upper branch computes attention weights for channel and spatial dimensions (C and W) using Z pooling followed by a convolutional layer and Sigmoid activation.
- The middle branch captures dependencies between channel dimensions (C) and spatial dimensions (H and W) using similar Z pooling and convolution operations followed by Sigmoid activation.
- The lower branch focuses on capturing dependencies among spatial dimensions (H and W) while maintaining the identity of the input, executing max pooling and convolution operations, followed by Sigmoid activation.
- After generating attention weights, each branch permutes the input, and their outputs are aggregated via average pooling to yield the final output of Triplet Attention.

This unique architecture of Triplet Attention, with its tailored operations across multiple dimensions and efficient computation, enables comprehensive feature representation and enhances a network's ability to discern critical features in various visual tasks. Additionally, its modular design facilitates seamless integration into existing network architectures, thereby augmenting their understanding and processing capabilities for complex data structures.

In summary, Triplet Attention stands out among other attention mechanisms in computer vision tasks due to its unique architecture and efficient computation. By leveraging a tripartite structure to capture cross-dimensional interactions, Triplet Attention effectively computes attention weights, enhancing a network's ability to focus on relevant features across different dimensions. This approach facilitates comprehensive feature representation and improves the model's performance in various visual tasks. Furthermore, Triplet Attention's modular design enables seamless integration into existing network architectures, augmenting their understanding and processing capabilities for complex data structures. This comparative analysis sets the stage for further exploration of Triplet Attention's effectiveness in the experimental section.

## 3. Methodology

*3.1. Focus on the spatial characteristics of receptive field*

The receptive field spatial feature refers to the local region of input data that the convolutional layer can "see" in the convolutional neural network[10] (CNN). In CNN, the output of each convolution operation is a small window, or a local receptive field, based on the input data. This receptive field defines the size and range of input data that the convolution kernel can access.

The concept of receptive fields is crucial to understanding how CNNS extract features from input data. In the primary layer of the network, the receptive field is usually small, allowing the model to capture subtle local features such as edges and corner points. As the data passes through more convolutional layers, the receptive field gradually expands by stacking layers on top of each other, allowing the network to perceive larger areas and capture more complex features such as textures and parts of objects.

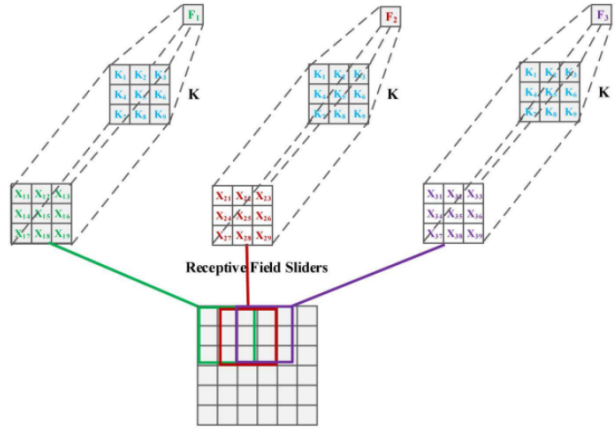

**Figure 4.** Convolutional network layer realizes k processing

The figure 4 above shows a 3x3 convolution operation. In this operation, the features are obtained by multiplying the convolution kernel with a receptive field slider of the same size and then summing it. Specifically, every 3x3 region (i.e. receptive field) on the input image X is processed by a 3x3 convolution kernel K. Each element in the receptive field, Xij(where and represents the position in the receptive field), is multiplied by the weight of the corresponding position in the convolution kernel K, Kij, and then these products are summed to give a new eigenvalue F. This process is carried out by sliding over the entire input image to generate a new feature map. This standard convolution operation emphasizes the concept of local join and weight sharing, i.e. the weight of the convolution kernel to the entire input graph.

In the context of CNN, receptive field spatial features refer to the features in the input image region that are perceived by each convolution operation. These features can include basic visual elements such as color, shape, and texture. In traditional convolutional networks, the receptive field is usually fixed, and each position is treated the same way. However, if the network can adapt the processing of the receptive field to the different characteristics of each region, then the network's understanding of the features will be more refined and adaptive.

Therefore, in this paper, we propose the RFAConv module, which aims to focus on the spatial characteristics of receptive fields. By introducing an adaptive receptive field mechanism, the RFAConv module can dynamically adjust the size and range of receptive fields according to local features of input data, thus capturing important spatial features more efficiently. Specifically, the RFAConv module utilizes an adaptive receptive field mechanism to weight the input data at each location to increase attention to important features, thereby enhancing the model's ability to understand complex scenes.

With the RFAConv module, we are able to extract key information from input data more accurately, providing a more reliable feature representation for subsequent target detection and recognition tasks.

*3.2. Mechanism to solve parameter sharing problem*

RFAConv convolution solves the problem of parameter sharing by introducing an attention mechanism that allows the network to assign a specific weight to each perception. In this way, the convolution kernel can dynamically adjust its parameters according to different features within each receptive field, rather than treating all regions equally.

Specifically, RFAConv uses spatial attention to determine the importance of each position in the receptive field and adjusts the weight of the convolution kernel accordingly. In this way, each receptive field has its own unique convolution kernel, rather than all receptive fields sharing the same kernel. This approach enables the network to learn local features in images in more detail, which helps improve overall network performance.

In this way, RFAConv improves the expressiveness of the model, allowing it to more accurately adapt and express the features of the input data, especially when dealing with complex or variable image content. [11-12]This dynamically adjusted parameter sharing mechanism enables the network to better capture the differences between different regions, thus enhancing the model's understanding and representation of image content.

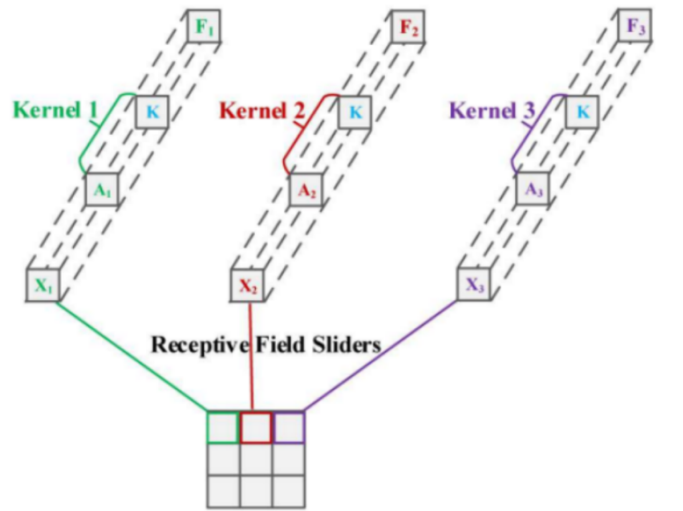

**Figure 5.** Schematic diagram of RFAConv convolution operation

During the RFAConv convolution operation, the convolution kernel parameter K is multiplied by the attention weight A to generate a customized convolution kernel for each receptor field position. Specifically, this process combines the attention mechanism with the convolutional kernel to produce a customized convolutional kernel for each receptive field location. For example, Kernel1, Kernel2, and Kernel3 in the figure are obtained by multiplying the common convolution kernel parameter K with the corresponding attention weights A1, A2, and A3, respectively. This method allows the network to assign different importance to the features of different spatial locations in the process of feature extraction, thus enhancing the ability of the model to capture key features.

Overall, such a mechanism increases the expressive power of convolutional neural networks, allowing the network to adapt more flexibly to different input characteristics and helping to improve the performance of the final task. This is an efficient way to deal with the problem of parameter sharing in traditional convolution operations because it allows the convolution kernel at each location to adapt to the specific region it is dealing with.

*3.3. Improve the efficiency of large-size convolution kernel*

RFAConv dynamically adjusts the weights of convolutional kernels by leveraging receptive field attention mechanism, providing customized attention for feature extraction in each region. This allows even large-sized convolutional kernels to effectively capture and process important spatial features without allocating excessive computational resources to less relevant information.

Specifically, the RFAConv method enables the network to identify and emphasize the more important regions in the input feature map and adjust the weights of convolutional kernels accordingly. This means that the network can reweight critical features, allowing large-sized convolutional kernels to not only capture a wide range of information but also concentrate computational resources on more informative features, thereby enhancing overall processing efficiency and network performance. This addresses the common phenomenon in standard convolution operations where feature overlap leads to weight sharing issues, implying that different receptive fields may use the same attention weights for the same input features.

In the illustration, F1, F2, FN represent the feature outputs within different receptive field sliders, obtained through element-wise multiplication of the input features X with corresponding attention weights A and convolutional kernel weights K. For instance, F1 is computed by multiplying X11 by the corresponding attention weight A11 and convolutional kernel weight K1, and so forth.

The diagram emphasizes that the parameters of convolution operations within each receptive field slider should not be entirely shared, but rather adjusted based on the features and corresponding attention weights in each specific region. This adjustment allows the network to handle each local region more finely, better capturing and responding to specific features of the input data rather than simply applying the same weights to the entire image. Such an approach enhances the network's understanding and representation of features, thereby improving learning and prediction outcomes.

In summary, through this approach, [13]RFAConv enhances the model's expressive power, allowing it to more accurately adapt to and represent the features of input data, especially when dealing with complex or variable image content. This flexible parameter adjustment mechanism provides a new pathway for improving the performance and generalization capability of convolutional neural networks.

### 4. Experimental Procedure

*4.1. Experimental design*

The main objective of this experiment is to improve the performance of YOLOv8 in target detection by improving its network structure. We use C2f_RFAConv to replace the C2f module in the original YOLOv8 network, and introduce RFAConv and Triplet Attention modules. We will verify the improvement effect by comparing the performance indicators before and after the improvement.

  1. Experimental environment:
- Hardware: GPU (e.g. NVIDIA Tesla V100)
- Software: Python 3.x, PyTorch, YOLOv8 framework
  2. Data set:
- Training and testing using the COCO dataset.

*4.2. Experimental Model*

  1.Original YOLOv8 network training:

Using the original YOLOv8 network structure, we trained on the COCO dataset, saving the model and log files during the training process. After the training, we recorded the final performance indicators with MAP(50) values of 0.326 and MAP(50-95) values of 0.187. In order to visually demonstrate the detection performance of the model, the precision rate-recall ratio (PR) curve of the original YOLOv8 was drawn and saved in this experiment.

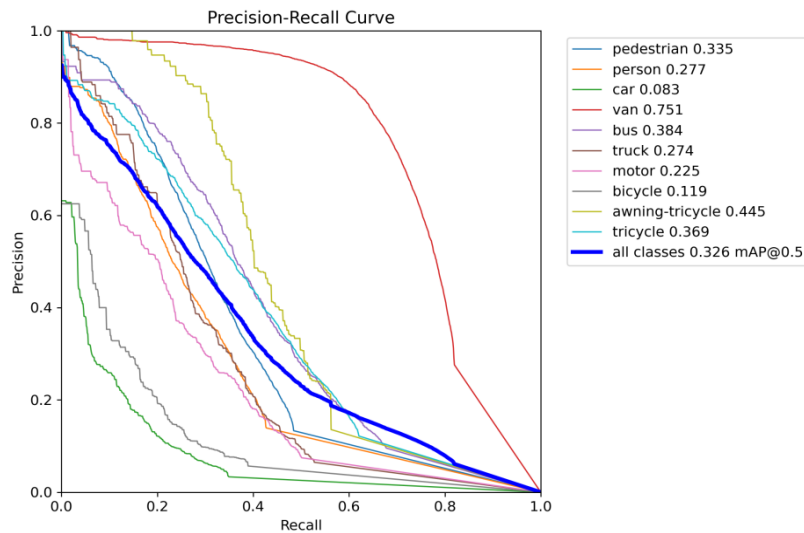

**Figure 6.** Accuracy - recall rate (PR) curve of the original YOLOv8

Figure PR curve shows the relationship between accuracy rate and recall rate of the model under different thresholds. The closer the curve is to the upper right corner, the better the performance of the model. In the PR curve, we can see the trend of the accuracy rate and recall rate of various targets with the detection threshold, which helps us to evaluate the performance of the model under different detection difficulties. Through these indicators and curves, we can fully understand the detection effect of the original YOLOv8 on the COCO data set, and provide a benchmark and reference for subsequent improvement.

2.Introduction of C2f_RFAConv module:

Replace C2f module in YOLOv8 network with C2f_RFAConv module to improve model performance. The concrete implementation steps include: First, we replace the Bottleneck structure in C2f with RFAConv, and then build a new C2f_RFAConv module. The implementation code for the C2f_RFAConv module is provided, which contains detailed definitions of the C2f_RFAConv and RFAConv classes. RFAConv makes convolution operations more efficient by introducing attention mechanisms and multi-scale feature extraction. We integrated these new modules into the YOLOv8 network to make sure the code was up and running. Through training on the COCO dataset, we verify the correctness and validity of the model.

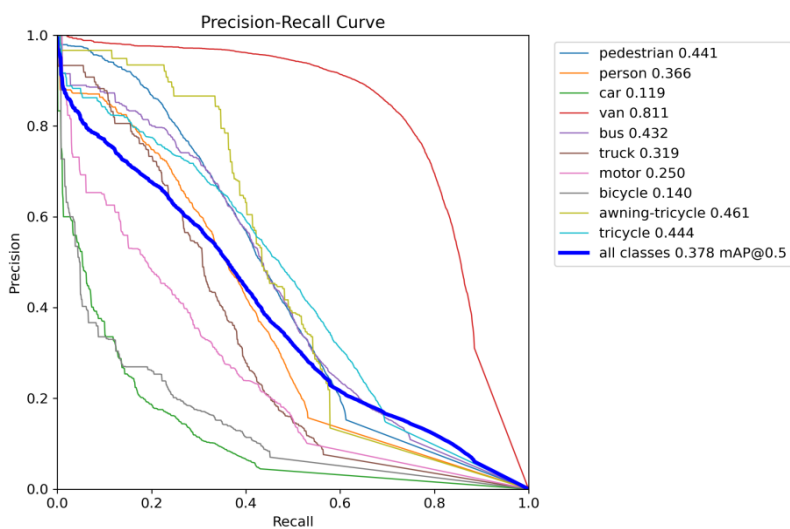

**Figure 7.** PR curve of the improved YOLOv8-C2f_RFAConv

After the completion of the experimental training, the improved performance indexes were recorded and a new PR curve was drawn. PR curve shows the relationship between accuracy rate and recall rate of the model under different thresholds. The curve should be closer to the upper right corner than the original model, indicating that the improved model has improved both accuracy and recall rate. The diagrams help us to fully evaluate the performance of the new model and provide the basis for subsequent optimization.

3.To introduce the Triplet Attention module:

The improved YOLOv8 network structure is used to train on the COCO dataset, and the training results are recorded, including the final MAP(50) value of 0.385 and MAP(50-95) value of 0.217. To visualize the performance of the improved network, we drew and saved a new PR curve. These curves show the relationship between accuracy rate and recall rate under different detection thresholds. The overall curve should be closer to the upper right corner than the original model, indicating that the improved model has significantly improved both accuracy and recall rate. These results and charts comprehensively evaluate the improved network performance and provide a basis for further optimization.

4.Performance comparison and analysis:

By comparing the performance indicators of the original YOLOv8 and the improved YOLOv8, we mainly focus on the changes of MAP(50) and MAP(50-95). The results show that the improved YOLOv8's MAP(50) has been improved from 0.326 to 0.385, and MAP(50-95) has been improved from 0.187 to 0.217. After calculating the performance improvement of each category, it is found that the detection accuracy of all categories has been improved in different degrees. The reasons for the performance improvement after the introduction of C2f_RFAConv and Triplet Attention modules are analyzed in detail. The main reason is that C2f_RFAConv module improves the expression capability and feature utilization of the network through more efficient feature extraction and weighting mechanism. The Triplet Attention module enhances the focus ability of the model in both spatial and channel dimensions, making the target detection more accurate. Together, these improvements contribute to a significant improvement in the overall performance of the model.

### 4.3. Experimental Result

- Original YOLOv8: MAP(50): 0.326-MAP(50-95): 0.187

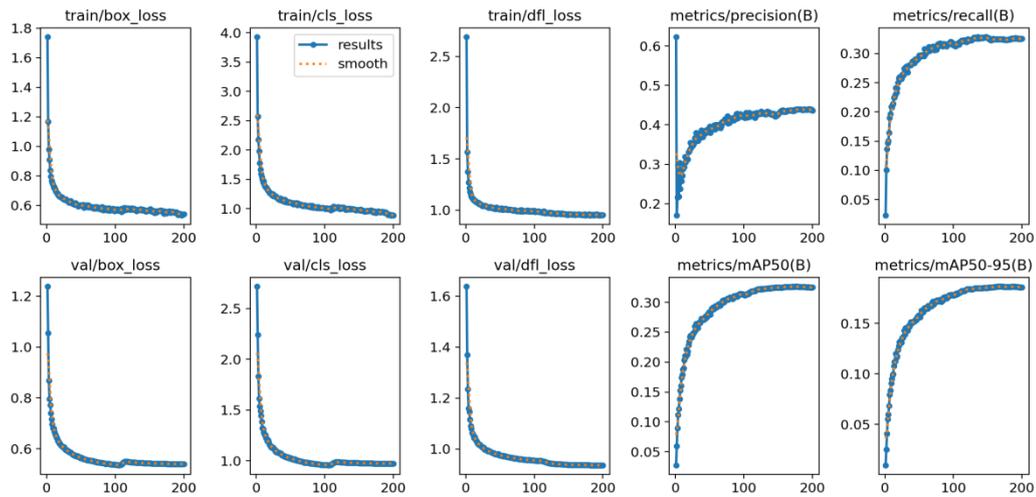

**Figure 9.** Training results of the original YOLOv8 model

- Improved YOLOv8-C2f_RFAConv-Triplet Attention-P2: MAP(50): 0.385-MAP(50-95): 0.217

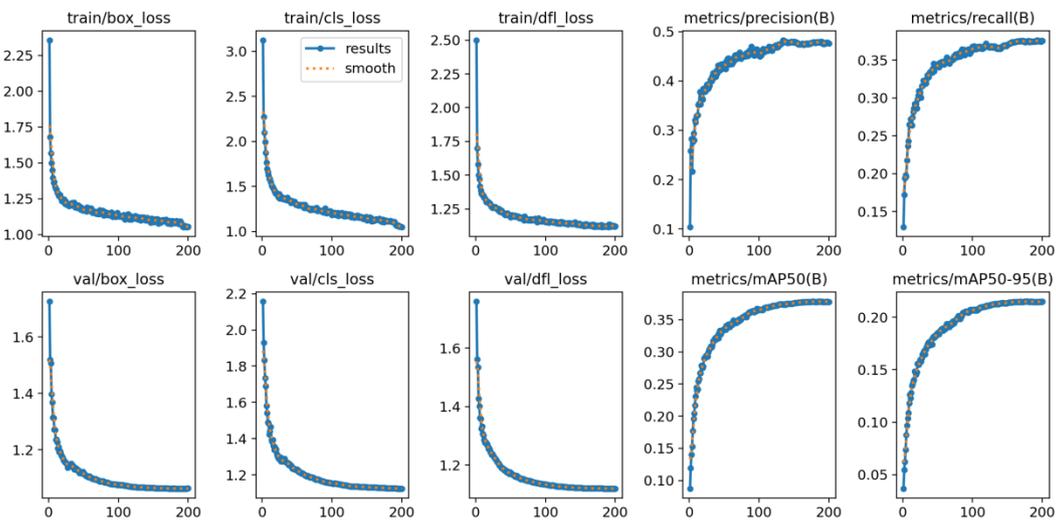

**Figure 10.** Result of improved YOLOv8-C2f_RFAConv-Triplet Attention-P2

Comparing the performance indicators of the original YOLOv8 and the improved YOLOV8-C2F_RfaconV-triplet Attention, the MAP(50) of the improved model has been improved from 0.326 to 0.385, with an increase of 0.059, and the MAP(50-95) has been improved from 0.187 to 0.217. An increase of 0.030. By analyzing the PR curve, the improved PR curve moves towards the upper right corner as a whole, showing that the accuracy rate and recall rate have been improved under various detection thresholds, indicating that the improved model has significantly improved the detection performance in each category. This trend further validates the effectiveness of introducing

C2f_RFAConv and Triplet Attention modules, which makes the model perform better in feature extraction and attention mechanism.

*4.4. Improvement Analysis*

In this experiment, the addition of detection head P2 significantly improved the detection performance of YOLOv8, especially in handling small targets, alleviating the impact of scale variance, enhancing robustness and reducing computation burden. P2 layer usually has higher resolution and is able to capture more details of small size targets, helping to improve small target detection by providing richer spatial information. At the same time, Layer P2 is located in the shallower layer of the network and can capture more fine-grained features, which is crucial for understanding the shape and texture of small targets. [14-15]This makes the detection ability of the model to small targets significantly improved. The detection head derived from layer P2 combined with the original detection head can effectively alleviate the negative impact of scale variance, so that the model has better adaptability and accuracy when dealing with different scale targets. In addition, P2 detection head has stronger anti-interference ability, can better adapt to different scenes and sizes of detection targets, improve the robustness of the algorithm.

In the improved analysis, the C2f_RFAConv module significantly improves the performance of the network through more efficient feature extraction and weighting mechanisms. C2f_RFAConv uses weighted feature extraction in RFAConv to capture important feature information more accurately and improve the expression ability of feature maps. The multi-scale feature extraction mechanism makes the model more flexible when dealing with objects of different sizes, and enhances the overall detection ability of the model.

The Triplet Attention module also plays an important role in feature extraction. By applying Attention mechanisms on both spatial and channel dimensions, Triplet Attention is able to focus more accurately on important feature areas, ignoring disturbing information. This dual attention mechanism greatly improves the accuracy of target detection in complex background. Attention in the spatial dimension can ensure that the model focuses on the local details of the target, while attention in the channel dimension optimizes the information interaction between the feature channels, thus improving the discriminant ability of the feature graph. In short, the combination of C2f_RFAConv and Triplet Attention module greatly enhances the feature extraction capability and detection accuracy of YOLOv8, making the model perform better in various detection tasks.

## 5. Conclusion

In conclusion, YOLOv8 stands as a cornerstone technology in the realm of autonomous driving, offering rapid and precise target detection capabilities vital for ensuring road safety and efficiency. This study has presented a comprehensive enhancement approach for the YOLOv8 model, addressing its crucial role and the challenges faced in autonomous driving scenarios. By introducing the C2f_RFAConv module and the Triplet Attention mechanism, significant improvements have been achieved in feature extraction efficiency and feature focusing, respectively. Through meticulous experimentation and evaluation, the effectiveness of these modifications has been demonstrated, as evidenced by notable increases in MAP values and enhancements in PR curves. Ultimately, the combination of these modifications results in a marked enhancement in the overall performance of the YOLOv8 model, promising improved safety and efficacy in autonomous driving systems.

In summary, the findings of this study underscore the pivotal role of advanced computer vision techniques, such as those integrated into the YOLOv8 model, in advancing the capabilities of autonomous driving systems. The successful implementation of the C2f_RFAConv module and the Triplet Attention mechanism has not only demonstrated the adaptability and versatility of the YOLOv8 framework but has also showcased its potential for further refinement and optimization. By leveraging these enhancements, the improved YOLOv8 model exhibits enhanced feature extraction efficiency and improved target detection accuracy, paving the way for safer and more efficient autonomous driving experiences in real-world scenarios.